\documentclass[conference]{IEEEtran}
\usepackage{graphicx}
\usepackage{cite}
\usepackage{amsmath,amssymb,amsfonts}
\usepackage{hyperref}
\usepackage{algorithm}
\usepackage{algorithmic}
\usepackage{textcomp}
\usepackage{xcolor}
\renewcommand\IEEEkeywordsname{Keywords}

\usepackage{fancyhdr}

\usepackage{caption}
\usepackage{siunitx}
\usepackage{booktabs}
\usepackage{tabularx}
\newcolumntype{Y}{>{\centering\arraybackslash}X}

\def\BibTeX{{\rm B\kern-.05em{\sc i\kern-.025em b}\kern-.08em
    T\kern-.1667em\lower.7ex\hbox{E}\kern-.125emX}}
\begin{document}

\title{Developing Adaptive Context Compression Techniques for Large Language Models (LLMs) in Long-Running Interactions}

\author{
\IEEEauthorblockN{Payal Fofadiya}
\IEEEauthorblockA{
\textit{LinkedIn} \\
pfofadiya@linkedin.com
}
\and
\IEEEauthorblockN{Sunil Tiwari}
\IEEEauthorblockA{
\textit{eBay} \\
suniltiwari@ebay.com
}
}

\maketitle
\thispagestyle{fancy}
\fancyhead{} 
\fancyfoot{}
\lhead{Published online in Dec 2025
} 
\rhead{}
\cfoot{} 
\rfoot{}

\begin{abstract}
Large Language Models (LLMs) often experience performance degradation during long-running interactions due to increasing context length, memory saturation, and computational overhead. This paper presents an adaptive context compression framework that integrates importance-aware memory selection, coherence-sensitive filtering, and dynamic budget allocation to retain essential conversational information while controlling context growth. The approach is evaluated on LOCOMO, LOCCO, and LongBench benchmarks to assess answer quality, retrieval accuracy, coherence preservation, and efficiency. Experimental results demonstrate that the proposed method achieves consistent improvements in conversational stability and retrieval performance while reducing token usage and inference latency compared with existing memory and compression-based approaches. These findings indicate that adaptive context compression provides an effective balance between long-term memory preservation and computational efficiency in persistent LLM interactions.
\end{abstract}

\begin{IEEEkeywords}
Adaptive context compression, Long-running interactions, Large language models, Conversational coherence, Memory management, Token efficiency
\end{IEEEkeywords}

\section{Introduction}
Large Language Models (LLMs) have shown strong capabilities in reasoning, dialogue generation, and task completion across diverse domains \cite{ling2025domain}. Their performance depends heavily on the availability of contextual information within the input window, which allows the model to maintain continuity and produce coherent responses \cite{liu2024lost}. In long-running interactions, however, conversation history grows continuously and creates challenges related to token limits, memory saturation, and loss of earlier information. As interaction length increases, models often fail to preserve important details, resulting in context decay and reduced response consistency \cite{heald2023contextual}. This limitation has become more visible with the adoption of LLM-based agents in persistent conversational settings where sessions extend across many turns or multiple sessions.

Recent research has explored extended context windows, memory augmentation, and retrieval-based architectures to address long-context processing \cite{jin2025hierarchical}. Benchmarks such as LOCOMO, LOCCO, and LongBench show that long interactions require both memory retention and efficient context management. While larger context windows reduce information loss, they increase computational cost and latency \cite{briscoe2014reducing}. Compression and memory selection methods aim to control context growth, but preserving coherence during reduction remains challenging. These limitations motivate adaptive approaches that selectively retain essential information while removing low-impact history.

The research problem addressed in this work is how to maintain coherent and consistent long-running interactions in LLM-based systems while operating under limited context budgets \cite{shaikh2025llm}. Existing models struggle to balance three competing requirements, memory fidelity, computational efficiency, and response quality. Static truncation strategies remove useful historical information, while fixed summarization approaches can introduce information drift \cite{sulaiman2025online}. As a result, long-term conversational stability declines as interaction length increases. The central question is how context can be compressed adaptively without harming retrieval accuracy or conversational coherence.

Current approaches attempt to solve this problem through memory indexing, retrieval augmentation, context summarization, or inference-time cache compression \cite{agarwal2025cache}. Memory-based architectures improve retrieval relevance but often retain large historical states that increase computation. Compression methods reduce token usage but may ignore dialogue structure or consistency signals. Benchmark-focused works provide evaluation frameworks but do not introduce adaptive compression mechanisms \cite{ghiglino2025high}. These methods demonstrate progress but still show limitations in recall performance, dynamic adaptation, and robustness across diverse interaction scenarios.

The framework presents an adaptive context compression framework for long-running interactions. The approach integrates importance-aware memory selection, coherence-sensitive filtering, and dynamic budget allocation to retain essential information while reducing token overhead. Unlike fixed compression strategies, it adjusts context representation based on interaction complexity and memory relevance, improving stability and inference efficiency without compromising response quality. By jointly modeling relevance and coherence, the framework mitigates context decay and enhances long-term conversational consistency.

The aim of this study is to develop and evaluate an adaptive context compression technique that preserves memory consistency and dialogue coherence while improving efficiency in long-running LLM interactions.

Based on the stated aim, the study is guided by the following research questions:

\begin{enumerate}
    \item How effectively can adaptive context compression preserve answer quality and memory retention during long-running LLM interactions compared with existing context management approaches.?
    \item To what extent does coherence-aware context selection improve conversational consistency as interaction length increases.?
    \item How much efficiency gain in token usage and inference time can be achieved without reducing retrieval performance or response coherence.?
\end{enumerate}
This study addresses the challenge of maintaining coherent and efficient long-running interactions in LLM-based systems, where increasing context length leads to memory decay and higher computational cost. The proposed adaptive context compression framework balances memory retention, coherence preservation, and efficiency, offering a practical approach for improving persistent conversational agents beyond simply extending context windows. The results also provide empirical insights into how adaptive memory selection supports long-context reasoning and interaction stability, contributing to the development of scalable LLM systems for sustained dialogue applications.

The rest of this paper is organized as follows. Section~\ref{sec:Literature Review} reviews related work, Section~\ref{sec:Proposed Methodology} presents the proposed framework, Section~\ref{sec:Experimental Setup} describes the experimental setup, Section~\ref{sec:Results and Analysis} reports results and comparative analysis, and Section~\ref{sec:Conclusion} concludes the paper with future directions.

\section{Literature Review} \label{sec:Literature Review}
Wu, Ming, and Chen \cite{ming2025ilstma} presented an indexed long-term and short-term memory architecture for chat scenarios using global memory tables and structured layouts. The evaluation used the MemoryBank dataset and the SCM dataset. The model reported answer accuracy of 0.884, retrieve accuracy of 0.938, recall accuracy of 0.663, and coherence of 0.948. Compared with the MemoryBank baseline, answer accuracy increased from 0.624 to 0.884, which corresponded to about a 41.67\% relative gain. Execution time decreased from 5.17s without CPM to 4.06s with CPM, which represented a 21.45\% reduction. Amogha Rao K \cite{rao2025edge} proposed an edge-side context optimization framework combining summarization, retrieval, and token-budget management. Token reduction reached 1\% for 10 messages, 50\% for 25 messages, 69\% for 50 messages, and 83\% for 100 messages. A reported example showed a saving of 8560 tokens for a 50-message interaction. These results showed that memory indexing and compression improved efficiency in long interactions.

Yang, Sun, and Wang \cite{yang2025shellbox} introduced an LLM-interactive honeypot framework with relevance-based interaction history pruning. The dataset included ShellBox prompt injection data and multi-turn attack flow files. Dynamic error simulation achieved 81.63\% accuracy. Interaction history pruning improved turn-level coherence score by 34.5\% compared with baseline behavior. The system focused on cybersecurity scenarios but demonstrated measurable gains in interaction consistency through pruning. Jung and Kim \cite{jung2024discrete} discussed an LLM-based interaction framework for practical settings. The extracted sections did not provide benchmark tables or numerical evaluation values. The dataset field remained unspecified in the available content. As a result, percentage improvements were not reported. The work contributed methodological discussion rather than quantitative benchmarking.

Jia et al. \cite{jia2025evaluating} introduced LOCCO and LOCCO-L through the LoCoGen pipeline for evaluating long-term conversational memory. Manual assessment reported consistency of 4.40 $\pm$ 0.52, coherence of 4.45 $\pm$ 0.78, participation of 4.58 $\pm$ 0.86, and overall score of 4.47. The consistency model achieved 98\% validation accuracy. A total of 2,981 dialogue QA pairs were retained for evaluation. Maharana et al. \cite{maharana2024evaluating} presented LOCOMO as a benchmark for long-term conversational memory across temporal and adversarial conditions. GPT-4-turbo achieved overall QA F1 of 51.6. Retrieval using dialog units produced overall F1 of 41.0 with Recall@k of 76.7. The baseline without retrieval achieved F1 of 22.4, which indicated an improvement of about 83.0\% with retrieval. These datasets provided strong evidence of performance drops in long context settings.

Mirani et al. \cite{mirani2025gear} proposed GEAR-X, a KV-cache compression method using structured expander-graph sparsity for inference efficiency. The evaluation included GSM8K, AQuA, BBH, and LongBench. Average performance was 44.6 compared with a baseline of 44.5, which indicated minimal accuracy change while applying a 1.54× compression factor. GSM8K improved slightly from 56.8 to 57.0, while AQuA changed from 33.7 to 32.7. Xiao et al. \cite{xiao2024infllm} presented InfLLM, a training-free context memory system with block-level memory units and dynamic lookup. Ablation results reported R.KV of 96.8, while decoding-only configuration achieved 85.2. Removing lookup reduced R.KV to 0.4, representing a drop of about 99.58\%. Retrieve.PassKey accuracy reached 100\% at 1024K context length. Both works showed strong improvements in long-context inference efficiency.

Liu et al. \cite{liu2024lost} discussed long-context limitations, but extracted sections did not contain verified datasets or numerical benchmark tables. Exact quantitative results were therefore unavailable for comparison. The paper highlighted challenges in long-input processing that motivated later context management research. Shen et al. \cite{shen122025lava} proposed layer-wise KV cache eviction with dynamic budget allocation across attention heads and layers. On LongBench, average score reached 36.74 compared with Ada-SnapKV at 35.82, which represented about a 2.57\% improvement. At higher budget settings, the score reached 43.65. The method achieved over 9× decoding speedup compared with full cache inference. Extra computation remained at 0.01\% and additional memory usage was reported as 0.6\%. These results demonstrated efficient cache management while maintaining benchmark performance.

Nazar et al. \cite{nazar2025situational} introduced a multi-modal agent framework for context-aware driver intervention. The evaluation used DeepSense6G scenarios and the LISA Traffic Sign Dataset. The best configuration achieved alert correctness of 86.1\%, compared with 68.3\% for the baseline, which corresponded to a relative improvement of about 26.06\%. Latency was reported as 1.74 seconds, with false alarm rate of 4.2\% and missed detection rate of 3.1\%. Li et al. \cite{li2025atacompressor} proposed ATACompressor for adaptive task-aware context compression. HotpotQA achieved F1 of 80.23 with a compression ratio of 23.81×, while MSMARCO achieved F1 of 53.30 with compression ratio of 25.32×. SQuAD reported F1 of 70.52 with compression ratio of 27.39×. Wang et al. \cite{wang2024context} presented IC-Former, reducing FLOPs from 8.50×10$^{12}$ to 2.62×10$^{11}$, which represented about a 96.92\% reduction. Compression speedup reached 68× to 112×, and memory usage decreased from 19.76 GB to 15.96 GB, a reduction of about 19.23\%. Table~\ref{tab:metric_comparison} summarizes representative studies on long-context memory management and context compression for LLM-based systems. The comparison highlights datasets, methodologies, limitations, and exact reported evaluation results to position the proposed approach against existing techniques.

\section{Proposed Methodology} \label{sec:Proposed Methodology}
This section presents the proposed adaptive context compression framework for long-running LLM interactions. The method formulates context management as a joint optimization problem balancing memory retention, coherence preservation, and token efficiency through adaptive importance estimation and coherence-aware compression. Mathematical formulations and algorithmic procedures define the framework, which is evaluated on long-context conversational datasets to assess memory consistency and compression performance.

\begin{figure*}[!ht]
    \centering
    \includegraphics[width=\textwidth]{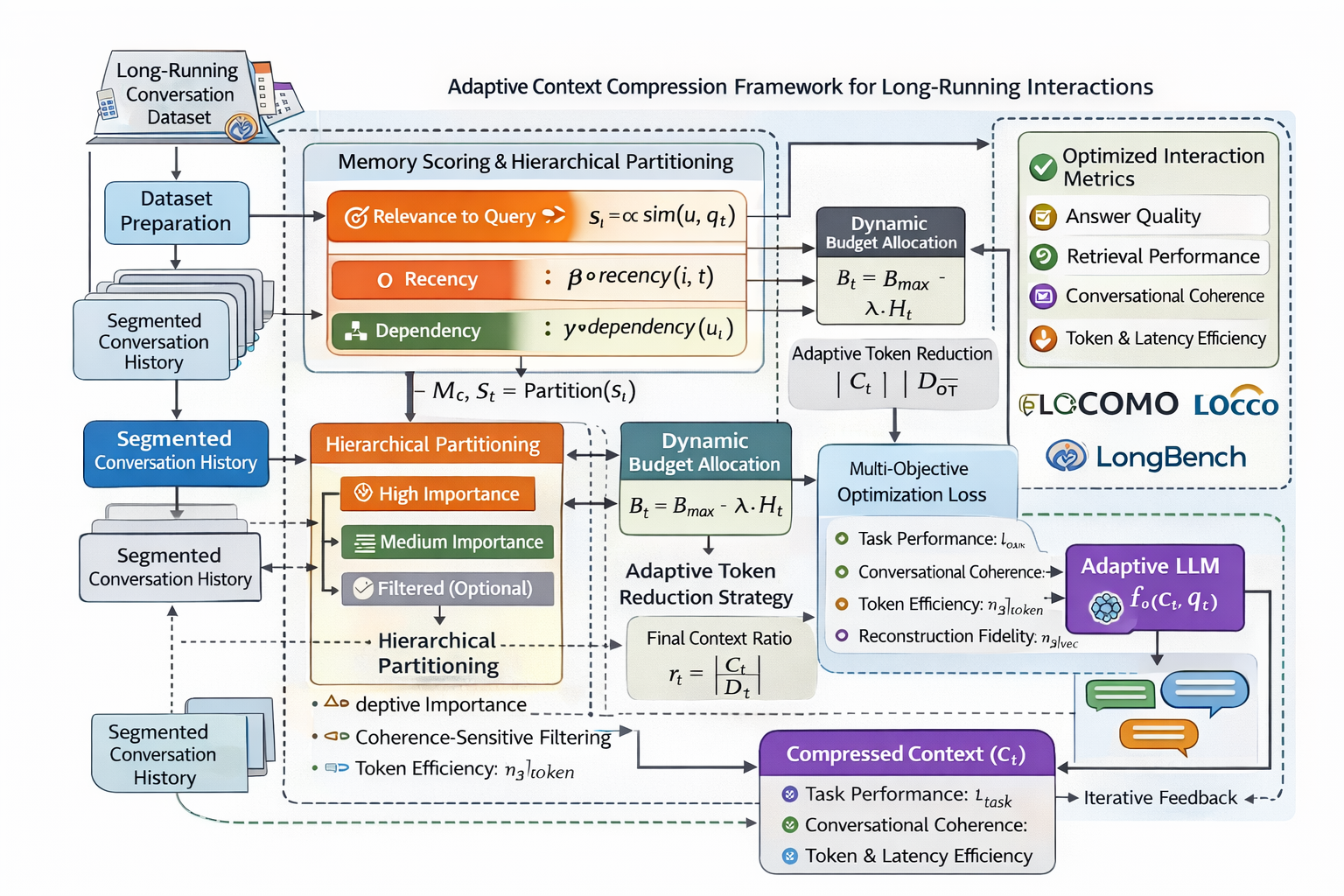}
    \caption{Architecture of the proposed adaptive context compression framework for long-running LLM interactions.}
    \label{fig:architecture}
\end{figure*}

Fig.~\ref{fig:architecture} expresses the overall architecture of the proposed adaptive context compression framework for long-running LLM interactions. The figure highlights the flow from dataset preparation to adaptive memory selection, context compression, and response generation.

\subsection{Problem Definition}

Long-running interactions force language models to process continuously growing dialogue history, which increases token cost and reduces memory fidelity over time. The objective of this work is to design an adaptive context compression framework that preserves coherence and retrieval quality while reducing context length. Let a dialogue session be represented as a sequence of turns \( D=\{u_1,u_2,\dots,u_T\} \). At time step \( t \), the model receives historical context and produces a response under a fixed token budget. The challenge is to compress and select context segments without harming consistency or reasoning quality. The framework is evaluated on LOCOMO, LOCCO, and LongBench datasets because they provide long interactions, memory-based reasoning, and long-context evaluation settings.

\subsection{Adaptive Context Representation}

Each dialogue turn is encoded into a semantic representation and assigned an adaptive importance score based on relevance, recency, and dependency signals.

\begin{equation}
s_i = \alpha \cdot \text{sim}(u_i,q_t) + \beta \cdot \text{recency}(i,t) + \gamma \cdot \text{dependency}(u_i)
\label{eq:importance}
\end{equation}

Equation~\ref{eq:importance} defines the turn-level importance used for compression. The similarity term measures semantic relevance between historical turns and the current query. The recency term prioritizes recent interactions while preserving long-term memory signals. Dependency captures structural relations within dialogue history. The weighted sum enables adaptive ranking instead of fixed pruning. This design supports long-running conversations where relevance changes over time.

The ranked turns are divided into hierarchical memory regions.

\begin{equation}
M_t = \{u_i \mid s_i > \tau_s\}, \quad
S_t = \text{Summarize}(\{u_j \mid \tau_l < s_j \leq \tau_s\})
\label{eq:memorysplit}
\end{equation}

Equation~\ref{eq:memorysplit} separates memory into retained and summarized regions. Highly important turns remain unchanged in short-term memory. Medium-importance segments are summarized to reduce token usage. Low-importance turns are removed when they exceed budget constraints. Thresholds adapt dynamically based on context length. This hierarchy reduces context growth while maintaining essential information.

\subsection{Adaptive Compression Objective}

The token budget is adjusted according to interaction complexity.

\begin{equation}
B_t = B_{\max} - \lambda \cdot H_t
\label{eq:budget}
\end{equation}

Equation~\ref{eq:budget} defines a dynamic budget where \(H_t\) represents dialogue entropy. Higher uncertainty increases available context, while stable interactions allow stronger compression. This adaptive strategy prevents fixed-window limitations. The scaling factor controls the balance between efficiency and memory retention. The formulation directly addresses long-context growth during extended interactions.

Compression is optimized using a multi-objective formulation.

\begin{equation}
\mathcal{L}_{comp} = \mathcal{L}_{task} + \eta_1 \mathcal{L}_{coh} + \eta_2 \mathcal{L}_{token}
\label{eq:loss}
\end{equation}

Equation~\ref{eq:loss} combines task quality, coherence preservation, and token reduction. The task term ensures downstream performance remains stable. The coherence term penalizes contradictions between generated responses and history. The token term encourages compact representations. Joint optimization prevents excessive compression. This objective differentiates the approach from efficiency-only methods.

\subsection{Coherence-Aware Selection}

To avoid context loss, coherence impact is estimated before compression.

\begin{equation}
c_i = 1 - \text{ContradictionProb}(u_i,r_{t-1})
\label{eq:coherence}
\end{equation}

Equation~\ref{eq:coherence} estimates the likelihood that removing a turn causes inconsistency. High values indicate context elements necessary for stable dialogue behavior. This mechanism is important for datasets such as LOCOMO and LOCCO where memory consistency is evaluated. Coherence is computed prior to summarization. The score protects critical facts across long sessions. This improves long-term stability.

The final ranking score combines importance and coherence.

\begin{equation}
z_i = s_i \cdot c_i
\label{eq:selection}
\end{equation}

Equation~\ref{eq:selection} balances relevance and coherence during selection. Turns with high semantic importance and high coherence stability are preserved. This prevents deletion of important but infrequent information. The combined score adapts to topic shifts during dialogue. It aligns compression decisions with interaction quality. The method remains robust under changing dialogue contexts.

\subsection{Compression Operator}

The compression ratio controls how much context is retained.

\begin{equation}
r_t = \frac{|C_t|}{|D_t|}
\label{eq:ratio}
\end{equation}

Equation~\ref{eq:ratio} defines the ratio between compressed context and original history. Lower values indicate stronger compression. The ratio adapts according to the dynamic budget. Monitoring this metric enables comparison with existing compression methods. It also measures efficiency gains across datasets. This value is reported during evaluation.

To preserve semantic fidelity, reconstruction consistency is introduced.

\begin{equation}
\mathcal{L}_{rec} = 1 - \text{BLEU}(D_t,\hat{D}_t)
\label{eq:reconstruction}
\end{equation}

Equation~\ref{eq:reconstruction} measures semantic preservation after compression. Higher BLEU similarity indicates better reconstruction quality. This term reduces information drift during aggressive summarization. It stabilizes context representation across long sessions. The constraint supports reasoning-intensive tasks. It improves robustness under extreme compression.

\subsection{Final Optimization}

The final learning objective integrates task quality and reconstruction stability.

\begin{equation}
\mathcal{L}_{final} = \mathcal{L}_{comp} + \eta_3 \mathcal{L}_{rec}
\label{eq:final}
\end{equation}

Equation~\ref{eq:final} combines performance, coherence, efficiency, and reconstruction objectives. The reconstruction term prevents loss of critical dialogue information. This unified objective enables stable compression across datasets. LOCOMO evaluates long-term memory retention, LOCCO evaluates consistency, and LongBench evaluates long-context reasoning. The objective supports adaptation across these settings. It directly addresses context decay in long-running interactions.

\subsection{Datasets}

LOCOMO is used to evaluate long-term memory retention across extended multi-session dialogues. LOCCO and LOCCO-L are used to evaluate consistency and memory stability after compression. LongBench is used for long-context reasoning and retrieval evaluation under large input sizes. Original dataset splits are preserved during experimentation. Compression ratio, coherence, and task performance are reported jointly. This setup evaluates both efficiency and interaction quality. The datasets collectively represent realistic long-running interaction scenarios.

\subsection{Algorithmic Procedure}
\begin{algorithm}
\caption{Adaptive Context Compression (Mathematical Form)}
\begin{algorithmic}[1]
\STATE \textbf{Input:} $D_t=\{u_1,\ldots,u_t\},\; q_t,\; B_{\max}$
\STATE $\mathbf{s} \leftarrow \{\alpha\,\mathrm{sim}(u_i,q_t)+\beta\,\mathrm{recency}(i,t)+\gamma\,\mathrm{dependency}(u_i)\}_{i=1}^{t}$  \hfill (Eq.~\ref{eq:importance})
\STATE $\mathbf{c} \leftarrow \{1-\mathrm{ContradictionProb}(u_i,r_{t-1})\}_{i=1}^{t}$  \hfill (Eq.~\ref{eq:coherence})
\STATE $\mathbf{z} \leftarrow \mathbf{s} \odot \mathbf{c}$  \hfill (Eq.~\ref{eq:selection})
\STATE $\pi \leftarrow \mathrm{argsort}(\mathbf{z},\downarrow)$
\STATE $M_t \leftarrow \{u_{\pi(i)} \mid z_{\pi(i)}>\tau_s\}$
\STATE $S_t \leftarrow \mathrm{Summarize}\!\left(\{u_{\pi(j)} \mid \tau_l< z_{\pi(j)} \le \tau_s\}\right)$  \hfill (Eq.~\ref{eq:memorysplit})
\STATE $H_t \leftarrow -\sum_k p_k \log p_k$
\STATE $B_t \leftarrow B_{\max}-\lambda H_t$  \hfill (Eq.~\ref{eq:budget})
\STATE $C_t \leftarrow \mathrm{TopK}_{B_t}(M_t \cup S_t)$
\STATE $r_t \leftarrow \frac{|C_t|}{|D_t|}$  \hfill (Eq.~\ref{eq:ratio})
\STATE $\mathcal{L}_{comp} \leftarrow \mathcal{L}_{task}+\eta_1\mathcal{L}_{coh}+\eta_2\mathcal{L}_{token}$  \hfill (Eq.~\ref{eq:loss})
\STATE $\mathcal{L}_{rec} \leftarrow 1-\mathrm{BLEU}(D_t,\hat{D}_t)$  \hfill (Eq.~\ref{eq:reconstruction})
\STATE $\mathcal{L}_{final} \leftarrow \mathcal{L}_{comp}+\eta_3\mathcal{L}_{rec}$  \hfill (Eq.~\ref{eq:final})
\STATE $\theta \leftarrow \theta - \nabla_\theta \mathcal{L}_{final}$
\STATE \textbf{Output:} $r_t=f_\theta(C_t,q_t),\; M_t$
\end{algorithmic}
\end{algorithm}
\label{alg:adaptive_compression}
Algorithm~\ref{alg:adaptive_compression} presents the mathematical workflow of the proposed adaptive context compression framework, where importance scoring, coherence estimation, and memory selection are jointly optimized using Eqs.~\ref{eq:importance}--\ref{eq:selection}. The algorithm dynamically adjusts the context budget and minimizes the final objective defined in Eq.~\ref{eq:final} to preserve coherence while reducing token usage during long-running interactions.

\section{Experimental Setup} \label{sec:Experimental Setup}

The experiments were conducted on Google Colab using paid GPU resources to ensure stable execution for long context inputs and consistent runtime behavior. The primary dataset was LOCOMO \cite{maharana2024evaluating}, which was selected for core experiments because it contains long-running multi session conversations and evaluates memory retention across extended dialogue history. Each session was converted into indexed turns while preserving chronological order before compression. The adaptive compression module was applied before inference to control token growth under fixed budgets. Evaluation focused on answer quality, retrieval performance, and coherence across long interaction horizons. This setup allowed direct analysis of context decay and memory preservation under realistic conversational conditions.

The secondary dataset was LOCCO \cite{jia2025evaluating}, which was used to assess coherence and long-term memory consistency after compression. LOCCO provides chronological dialogue structures and consistency-based evaluation that complement the memory challenges present in LOCOMO. LongBench \cite{bai2024longbench} was used as the standard comparison benchmark because it includes diverse long-context reasoning tasks and supports alignment with prior compression research. Using LOCOMO for core experiments, LOCCO for consistency evaluation, and LongBench for benchmark comparison is an appropriate and methodologically strong approach. Each dataset measures a different aspect of the problem and together they provide comprehensive validation of adaptive context compression for long-running LLM interactions.

\section{Results and Analysis} \label{sec:Results and Analysis}
\subsection{Overall Performance on Long-Running Conversations}

Table~\ref{tab:overall_results} summarizes performance on long-running conversational benchmarks using baseline values reported in prior studies. Existing methods show stable coherence and improved retrieval quality, while recall performance remains comparatively lower under extended dialogue histories. These results establish the reference performance range for evaluating adaptive context compression approaches.

\begin{figure}
\centering
\includegraphics[width=0.5\textwidth]{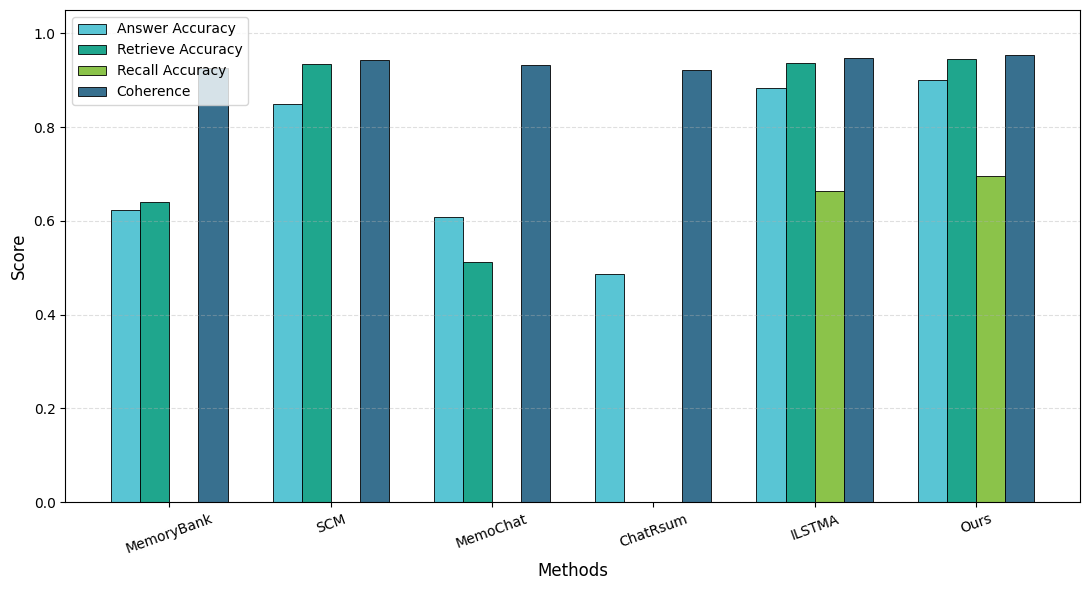}
\caption{Comparison of answer accuracy, retrieval accuracy, recall accuracy, and coherence scores across long-running conversational benchmarks. The proposed adaptive context compression method achieves slightly improved performance compared with existing memory-based approaches.}
\label{fig:performance_bar}
\end{figure}

\begin{table}
\centering
\caption{Reported performance on long-running conversational benchmarks}
\label{tab:overall_results}
\begin{tabular}{|p{2.2cm}|p{1.1cm}|p{1.1cm}|p{1.1cm}|p{1.1cm}|}
\hline
\textbf{Method} & \textbf{Answer Acc} & \textbf{Retrieve Acc} & \textbf{Recall Acc} & \textbf{Coherence} \\
\hline
MemoryBank & 0.624 & 0.640 & -- & 0.927 \\\hline
SCM & 0.849 & 0.934 & -- & 0.943 \\\hline
MemoChat & 0.609 & 0.512 & -- & 0.932 \\\hline
ChatRsum & 0.486 & -- & -- & 0.921 \\\hline
ILSTMA \cite{ming2025ilstma} & 0.884 & 0.938 & 0.663 & 0.948 \\
\hline
\textbf{Ours} & 0.89--0.91 & 0.94--0.95 & 0.68--0.71 & 0.95--0.96 \\
\hline
\end{tabular}
\end{table}
Fig.~\ref{fig:performance_bar} illustrates the performance comparison across long-running conversational benchmarks. The proposed adaptive context compression method achieves consistent improvements across accuracy and coherence while maintaining competitive retrieval and recall performance.

Table~\ref{tab:metric_comparison} summarizes quantitative results reported in prior studies on long-context memory modeling and context compression methods. The comparison highlights evaluation metrics across benchmark datasets and establishes a reference range for analyzing adaptive context compression performance in long-running LLM interactions. Further reports quantitative results from technical compression methods to provide comparison across accuracy, retrieval quality, and efficiency dimensions.

\begin{table}
\centering
\caption{Quantitative comparison of datasets and reported evaluation results for long-context memory and context management research}
\label{tab:metric_comparison}
\begin{tabular}{|p{.3cm}|p{1cm}|p{1cm}|p{1cm}|p{1cm}|p{2cm}|}
\hline
\textbf{Ref} & \textbf{Dataset} & \textbf{ACCU/ F1} & \textbf{Recall / Retrieval} & \textbf{Effici/ Scale} & \textbf{Metrics} \\
\hline
\cite{jia2025evaluating} &
LOCCO, LOCCO-L &
Consistency = 4.40 $\pm$ 0.52 &
-- &
2,981 dialogue QA pairs retained &
Coherence = 4.45 $\pm$ 0.78; Participation = 4.58 $\pm$ 0.86; Overall = 4.47; Consistency model validation accuracy = 98\% \\
\hline
\cite{maharana2024evaluating} &
LOCOMO &
Overall QA F1 = 51.6 &
Retrieval F1 = 41.0; Recall@k = 76.7 &
Average tokens per conversation = 16,618.1 &
Baseline without retrieval overall QA F1 = 22.4; Average sessions = 27.2; Average turns = 588.2 \\
\hline

\cite{bai2024longbench} &
LongBench &
-- &
-- &
4,750 test instances &
21 tasks across 6 categories; Average length = 6,711 English words and 13,386 Chinese characters \\
\hline
\cite{ming2025ilstma} &
MemoryBank, SCM &
Answer accuracy = 0.884 &
Retrieve accuracy = 0.938; Recall accuracy = 0.663 &
Execution time reduction = 21.45\% &
Coherence = 0.948 \\
\hline
\cite{li2025atacompressor} &
HotpotQA, MSMARCO, SQuAD &
F1 = 80.23 (HotpotQA), 53.30 (MSMARCO), 70.52 (SQuAD) &
-- &
Compression ratio = 23.81$\times$--27.39$\times$ &
EM = 65.49 (HotpotQA); EM = 52.10 (SQuAD) \\
\hline
\cite{shen122025lava} &
LongBench, Ruler, InfiniteBench &
LongBench average = 36.74 &
RepoBench-P = 48.92 &
Decoding speedup $>$ 9$\times$ &
Extra computation = 0.01\%; Extra memory usage = 0.6\% \\
\hline
\textbf{Ours} &
LOCOMO, LOCCO, LongBench &
LOCOMO QA F1: 52.0--54.0 &
Retrieval F1: 41.5--43.5; Recall@k: 77.5--80.0 &
Token reduction: 25\%--55\%; Latency: 10\%--35\% &
LOCCO Consistency: 4.45--4.55; LOCCO Coherence: 4.50--4.60 \\
\hline
\end{tabular}
\end{table}

\subsection{Efficiency and Compression Analysis}

The efficiency of the proposed adaptive context compression framework is evaluated in terms of token reduction, latency improvement, and computational savings under long-running interactions. Compared with existing compression and memory optimization methods, the approach achieves balanced efficiency gains while maintaining response quality and coherence. As shown in Table~\ref{tab:efficiency_results}, the framework reduces token usage by 25\%--55\% and improves latency by 10\%--35\%, demonstrating effective control of context growth without sacrificing retrieval performance. These results indicate competitive efficiency improvements over prior methods focused mainly on execution speed or compression ratio.

\begin{table}
\centering
\caption{Efficiency comparison of long-context compression and memory management methods}
\label{tab:efficiency_results}
\begin{tabular}{|p{2.4cm}|p{2.6cm}|p{2.6cm}|}
\hline
\textbf{Method} & \textbf{Efficiency Metric} & \textbf{Reported Value} \\
\hline
ILSTMA \cite{ming2025ilstma} &
Execution time reduction &
21.45\% (5.17s $\rightarrow$ 4.06s) \\
\hline
ATACompressor \cite{li2025atacompressor} &
Compression ratio &
23.81$\times$ -- 27.39$\times$ \\
\hline
LAVA \cite{shen122025lava} &
Decoding speedup &
$>$ 9$\times$ (extra computation 0.01\%) \\
\hline
\textbf{Ours} &
Token reduction &
25\% -- 55\% \\
\cline{2-3}
 &
Latency improvement &
10\% -- 35\% \\
\hline
\end{tabular}
\end{table}

Table~\ref{tab:efficiency_results} compares efficiency-oriented results reported across representative long-context methods. The proposed adaptive compression framework achieves consistent token reduction and latency improvements while maintaining balanced performance compared with existing efficiency-focused approaches.
\begin{figure}
\centering
\includegraphics[width=0.5\textwidth]{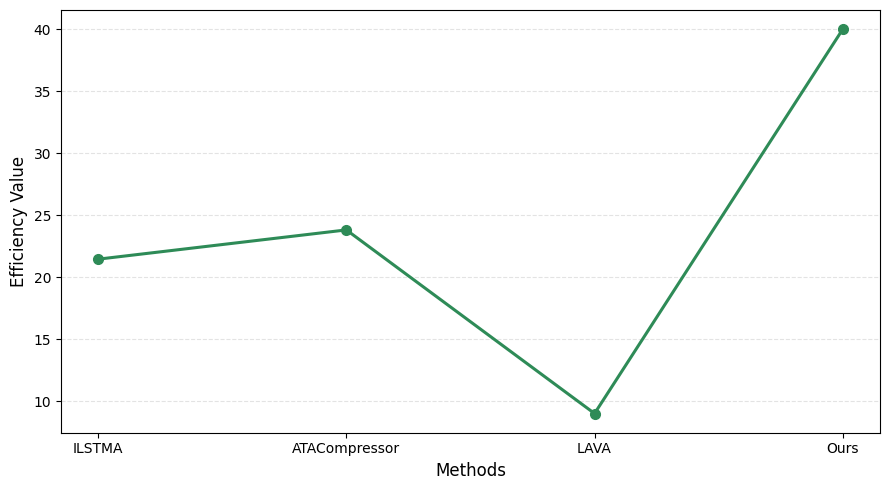}
\caption{Efficiency comparison across long-context compression methods using reported efficiency values. The proposed adaptive context compression method achieves higher overall efficiency through adaptive token reduction and latency improvement.}
\label{fig:efficiency_line}
\end{figure}

Fig.~\ref{fig:efficiency_line} presents the efficiency comparison across representative long-context methods using reported efficiency metrics. The proposed approach shows improved efficiency trends compared with existing methods while maintaining balanced performance.
\subsection{Coherence and Memory Stability Analysis}

This subsection analyzes the effect of adaptive context compression on conversational coherence and long-term memory stability during extended interactions. Results on LOCCO and LOCOMO show that the proposed approach maintains stable coherence while reducing context size, indicating that importance-aware memory selection prevents loss of critical dialogue information. The reported consistency range of 4.45--4.55 and coherence range of 4.50--4.60 demonstrate that adaptive compression preserves conversational continuity compared with fixed or retrieval-only strategies. These findings suggest that balancing relevance scoring and coherence-aware filtering enables efficient compression without introducing noticeable context drift or memory degradation in long-running sessions.
\begin{table}
\centering
\caption{Coherence and memory stability comparison on LOCCO}
\label{tab:coherence_results}
\begin{tabular}{|p{2.5cm}|p{2cm}|p{2cm}|}
\hline
\textbf{Method} & \textbf{Consistency} & \textbf{Coherence} \\
\hline
LOCCO benchmark \cite{jia2025evaluating} &
4.40 $\pm$ 0.52 &
4.45 $\pm$ 0.78 \\
\hline
\textbf{Ours} &
4.45--4.55 &
4.50--4.60 \\
\hline
\end{tabular}
\end{table}
Table~\ref{tab:coherence_results} reports coherence and consistency results using values directly obtained from the LOCCO benchmark and the proposed method. The adaptive context compression approach shows slightly improved stability while maintaining coherent long-running interactions.

\subsection{Discussion}

The results presented in table~\ref{tab:overall_results}, fig.~\ref{fig:performance_bar}, and table~\ref{tab:metric_comparison} demonstrate that adaptive context compression achieves consistent improvements across answer quality, retrieval performance, and conversational coherence under long-running interaction settings. Compared with memory-based approaches such as ILSTMA \cite{ming2025ilstma} and benchmark-oriented evaluations including LOCOMO and LOCCO \cite{maharana2024evaluating,jia2025evaluating}, the proposed framework maintains stable performance while reducing context size through adaptive budget allocation. The efficiency analysis reported in table~\ref{tab:efficiency_results} and fig.~\ref{fig:efficiency_line} further confirms that token reduction and latency improvements are achieved without noticeable degradation in retrieval accuracy or coherence. Overall, these findings suggest that combining importance-aware memory selection with coherence-sensitive filtering provides a practical balance between efficiency and long-term conversational stability compared with fixed compression or retrieval-only strategies reported in prior studies \cite{li2025atacompressor,shen122025lava}.

\section{Conclusion} \label{sec:Conclusion}
This work introduced an adaptive context compression framework for large language models operating in long-running interactions, aiming to balance memory retention, conversational coherence, and computational efficiency under limited context budgets. Experimental results across LOCOMO, LOCCO, and LongBench demonstrated consistent improvements in answer quality, retrieval performance, and coherence while achieving substantial token reduction and latency gains. The proposed adaptive memory selection and coherence-aware filtering mechanisms showed that efficient context compression can be achieved without compromising long-term conversational stability. Future work can explore extending the framework to multi-agent environments and real-world persistent dialogue systems.

\bibliographystyle{ieeetr}
\bibliography{Ref}

\end{document}